\documentclass{article}

\usepackage[preprint,nonatbib]{neurips_2020}

\usepackage[utf8]{inputenc} % allow utf-8 input
\usepackage[T1]{fontenc}    % use 8-bit T1 fonts
\usepackage{hyperref}       % hyperlinks
\usepackage{url}            % simple URL typesetting
\usepackage{booktabs}       % professional-quality tables
\usepackage{amsfonts}       % blackboard math symbols
\usepackage{nicefrac}       % compact symbols for 1/2, etc.
\usepackage{microtype}      % microtypography
\usepackage{graphicx}
\usepackage{caption}
\graphicspath{ {figures/} }
\usepackage{titlesec}
\title{Grading video interviews with fairness considerations}

\author{%
  Abhishek Singhania \\
  Aspiring Minds, A SHL Company\\
  \texttt{abhishek.singhania@aspiringminds.com} \\
   \And
   Abhishek Unnam \\
   Aspiring Minds, A SHL Company\\
   \texttt{abhishek.unnam@aspiringminds.com} \\
   \And
   Varun Aggarwal \\
   Aspiring Minds, A SHL Company\\
   \texttt{varun@aspiringminds.com} \\
}
\begin{document}
\maketitle
\begin{abstract}
There has been considerable interest in predicting human emotions and traits using facial images and videos. Lately, such work has come under criticism for poor labeling practices, inconclusive prediction results and fairness considerations. We present a careful methodology to automatically derive social skills of candidates based on their video response to interview questions. We, for the first time, include video data from multiple countries encompassing multiple ethnicities. Also, the videos were rated by individuals from multiple racial backgrounds, following several best practices, to achieve a consensus and unbiased measure of social skills. We develop two machine-learning models to predict social skills. The first model employs expert-guidance to use plausibly causal features. The second uses deep learning and depends solely on the empirical correlations present in the data. We compare errors of both these models, study the specificity of the models and make recommendations. We further analyze fairness by studying the errors of models by race and gender. We verify the usefulness of our models by determining how well they predict interview outcomes for candidates. Overall, the study provides strong support for using artificial intelligence for video interview scoring, while taking care of fairness and ethical considerations.
\end{abstract}
\section{Introduction}
The recruitment process at companies and organizations generally consists of assessments and personal interviews. Assessments consisting of multiple-choice tests were automated way back in the 1990s. Recently there have been efforts to automate the scoring of open-response tests , such as those to measure spoken English \cite{shashidhar2015acl,shashidhar2015kdd}, programming skills \cite{srikant2014system,takhar2019grading,Aggarwal:2016:ADC:2888451.2892037}, essay and email writing skills \cite{unnamgrading}. Over the last few years, advancements in computer vision have spurred interest in the grading video interviews \cite{hemamou2019hirenet,hireme,chen2017automated} automatically.

There have been several research studies \cite{padgett1995identifying,kohler2004differences,bargal2016emotion} and proliferation of tools \cite{mcduff2016affdex,amazonemotion,azureemotion} that attempt to score emotions based on images/videos. In \cite{chen2017automated}, the authors develop algorithms to predict personality based on video interviews, while in \cite{hemamou2019hirenet}, the authors develop models to predict hirability based on job profiles and video interview feeds. In \cite{naim2016automated}, the authors record actual interviews in a lab setting and then develop models to predict social skills.

However, such research has attracted criticism, particularly from the psychology community, over considerations of ethics and fairness. In \cite{doi:10.1177/1529100619832930}, the authors question the very hypothesis that emotions, that are internal states, could be rated based on external behaviors, such as facial expressions and voice characteristics. They also question the process of labeling emotions, the universality of emotional manifestation across cultures and the inadequate criteria being used to judge the quality of models. Furthermore, variations in the results of image and video processing with regard to gender, race, nationality, etc., have raised questions of fairness \cite{buolamwini2018gender, muthukumar2019color, cook2019demographic}. Algorithms tend to pick non-causal markers \cite{obermeyer2019dissecting,causalDisc} that signal one’s identity than behavior. On the other hand, techniques that predict hireability status propagate bias in extant hiring practices. It is unclear which skills and behaviors these model actually predict and they are susceptible to using spurious correlations in the sample. There has been no study on fairness of video interview grading algorithms, other than \cite{ETSFairnessVideo}, which looks at biases in the labelling process. Grading of video interviews has enormous implications for a person’s economic outcomes and living standard. Therefore, the bar for fairness and ethics must be set high for grading video interviews. In this paper, we mitigate these criticisms for the first time and explain how we developed models to grade video interviews with fairness considerations. We take several steps to do so.

We, for the first time, use a multi-country, multi-racial dataset to develop video-interview grading. Our data of 810 candidates and 5845 videos includes people from the US, UK, and India, who identify as Caucasians, African Americans, Asian, Hispanic\footnote{We merge Hispanic with few smaller ethnic groups and call them "Others".} and a few other smaller ethnic groups. We apply significant thought to devise meaningful parameters for rating video interviews and to establish a fair process to get the labels. Rather than attempting to gauge internal states, we measure externally expressed behavior. We refer to such behavior as “social skills” that encompass parameters such as “Confidence” (e.g., whether the average person watching the video would agree that the candidate projects confidence). Each parameter includes a proper rubric along with a ‘Cannot say’ option if the rater decides that the information in the video is not sufficient to grade a given parameter\footnote{In addition to social skills, we also measure interviews according to the spoken content. However, that is not the subject of this paper.}. The video interviews are rated for social skills by multiple raters of different racial backgrounds. The scores of these various raters correlate moderately well. We consider their consensus rating, to comprise the universal understanding of social skills.

We develop two models to predict social skills. The first uses expert-guidance to pick causal features. We further constrain the modeling space to avoid the use of spurious non-causal features\footnote{The model is not necessarily causal, but has been guided in that direction.}. Our intention is to build a relatively simple model with well-meaning features, that is interpretable and may show less variance over different samples. In the second model, we give machine learning a free hand to exploit the correlations in the dataset to build the best possible predictive model. Here, we use a state-of-the-art transformer model with additive attention to predict the grades. We recommend that if a simpler, causal-guided model gives comparable accuracy to a more expressive model, the former is preferable out of consideration of fairness. In fact, we discover that the simpler model performs as well as the deep-learning model on two parameters, while the deep-learning based model performs better on the third.

We examine further whether our models are biased and compare model errors by race and gender. We find that effect sizes are small in error for all parameters barring one involving race, where also it is borderline. We discuss methods to mitigate this. Finally, we test our social skill predictions against interview outcomes, using dataset from three companies. We find that social skills are indeed consistent predictors of interview outcomes.

This work makes significant new contributions:
\begin{itemize}
    \setlength\itemsep{-0.1em}
    \item It is the first video grading work to look at fairness with regard to gender and race.
    \item We introduce and use the first multi-country and multi-racial dataset of video interviews.
    \item We introduce a new approach to looking at building fair models: comparing simpler expert-guided models with complex fully-empirical models.
    \item We for the first time study the errors of video grading models with regard to gender and race.
\end{itemize}
We believe this work will go a long way in setting context and methods for fairness studies in video processing in general and video interviews in particular.
\section{Dataset and Ratings}\label{dataset_ratings}
\subsection{Rating Parameters}
Traditionally, researchers process facial images to predict the candidate’s emotions (such as happy, sad, anger, etc.). They use facial action units, (i.e., facial expressions) and map them to emotions \cite{keltner2003facial}. However, such research has been subject to criticism for several reasons. For example, emotions are purported to be internal states that are not necessarily signaled by facial expressions alone. Furthermore, the experience and perception of emotions may vary across cultures, races and countries. Also, images alone may not be sufficient to make a determination of a person’s emotional state. Nevertheless, raters are often implicitly forced to assign labels even if they think that the information is insufficient to make a confident designation.

In our approach, we take several steps to address these criticisms. First, we rate videos instead of still images. The videos have a duration of at least 30 seconds. This gives the raters ample time to observe the candidates before making a judgement. Moreover, a candidate’s final rating is based on a set of videos (at least six) rather than a single video. Also, our rating parameters are expressed behaviors rather than internal states. Several studies show that expressed behavior through facial expressions and voice tone, are judged during interviews and correlate to performance. \cite{nvb} finds significant difference in non-verbal behavior such as facial expressions, head movements and eye-contact between selected and rejected candidates in an interview. \cite{teachingL2} lists enthusiasm, tempo, and body language as pragmatic skills needed in a interview, and \cite{visualcues} finds visual cues such as smile, gaze and body movement are predictive of on-job performance.

We finalized on four rating parameters - Positive Emotion, Calmness, Confidence and Engagement. We call such expressed behavior “social skills.” These skills are essential in such roles as sales, customer service and management. There is a rubric related with each parameter, and the rater assigns a score by selecting a level on the rubric. The first option (in bold for emphasis) on the rubric indicates that the video cannot be rated on the parameter. This option is provided in order to eliminate any parameter for which the video offers insufficient information. Refer to supplementary material for the rubric used for rating.

As we will discuss in further sections, the candidates and raters in our study span diverse cultures and racial backgrounds. This was done to a universal, or stated scientifically, consensus (shared-view), rating of social skills.
\subsection{Video Dataset}
Each candidate must answer seven pre-recorded questions in English. The candidate is allotted 30 seconds to think about her/his response, and then one minute to deliver the response. The sessions are recorded via the candidate’s own webcam. The questions include situational questions, competency-based questions and domain knowledge questions from areas such as technology, banking, accounting, etc. (See examples in the supplementary material).

We collected data from 810 jobseekers from the US, UK, India and some other countries from Europe. We designed the sample to include candidates of different age, gender, racial and educational background. We collected a total of 5845 videos (with duration of thirty seconds to a minute) from the 810 jobseekers. The diversity of the dataset allows us to test whether our algorithms are biased against any of the various groups and how one may mitigate such bias. To the best of our knowledge, this study is the first to include data from multiple countries and multiple ethnic groups to measure social skills, emotions or grade video interviews. Refer to supplementary material to view the distribution of candidates by country, age, gender and race.   
\subsection{Raters and rating process}
Many previous studies have examined whether the ML algorithm exhibits bias towards any particular group. In \cite{ETSFairnessVideo} authors examine the prospect of human bias in scoring video-based structured interviews. The possibility of bias in labeling takes on increasing importance in the present case for two reasons. First, we are rating faces. Uncorrelated facial markers such as race, color, gender, etc. may bias the raters. Second, social skills may be perceived differently by individuals of different gender and racial backgrounds. We aim to capture the common variance (consensus) among these different groups to establish a universal social skills scoring index.

We recruited our raters online. All raters have experience working as HR recruiters, soft skills trainers, or possessed a background in industrial organizational psychology. We chose the rater sample to be representative of multiple ethnicities and gender. We performed a first rater selection exercise. Here, each rater scored 50 videos. We removed the raters with an average inter-rater correlation of less than 0.5 and a mean-difference of more than 1.0 (on a 5-point scale). This produced a pool of 31 raters, from diverse backgrounds. This exercise also helped us determine how many raters are required per video to provide a stable consensus rating. We bootstrapped different numbers of raters and studied how the variance in consensus ratings decreased with every additional rater. We observe little reduction in score variance when adding additional raters beyond five.

We then had every video rated by 7 raters. We again removed the raters which didn’t agree with others, on average (same criteria as before). Finally, each video was rated by atleast 5 raters. The distribution of the raters on race and gender is provided in supplementary material. We used the mean ratings of raters, as the final rating per video, per social skill.
\section{Methods}
We develop video-wise models for each social skill using supervised learning. We derive a candidate-level score by averaging the video-wise scores. We present two approaches to develop models. In the first approach, we solicit expert advice in choosing theoretically valid features and also constrain the model-space through certain computational techniques. The idea is to develop a model that is less susceptible to non-causal correlations and sampling biases. In the second approach, we use a much more expressive Deep Learning model that relies solely on the correlations present in the sample dataset. One may refer to supplementary material for the related work in the field of automated grading of video-interviews. We wish to determine whether the Deep Learning approach results in greater accuracy, and if so, how much. If there is no significant difference in model accuracy, then we prefer the first approach given that it provides a better theoretical basis and may better generalize over different kinds of samples. We now describe the two methods we use.
\subsection{Expert-Driven Approach}
\begin{figure}
    \centering
    \fbox{
    \includegraphics[width=0.95\textwidth]{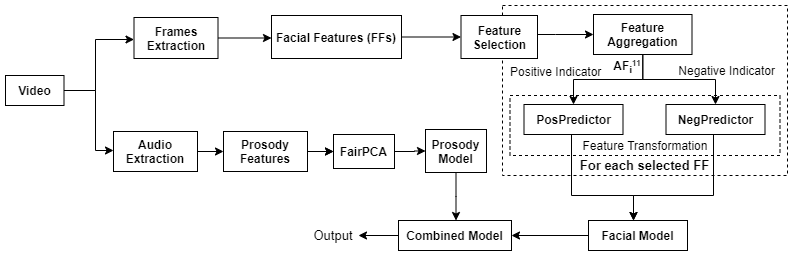}
    }
    \caption{Expert-Driven approach flow diagram for a social skill.}
    \label{mlflow}
    \vspace{-4mm}
\end{figure}

In this approach, we derive several features from the video, select certain features based on expert guidance, make some feature transformations and finally use a classical supervised learning technique to train models against the ratings. All these steps are illustrated in Fig \ref{mlflow}. We first describe the features that we used.
\subsubsection{Feature Engineering}
We primarily use two kind of features: 

\textbf{Facial Features (FFs):} We first extract the frames from the video at a sampling frequency of 15 frames per second. We use OpenFace \cite{7477553} to derive the intensities of $17$ different facial action units (AU) from each frame. Facial action units comprise facial movements such as lip curl, and eye brows raise. We also derive $6$ head pose translations and rotations (HP). This results in a $23$-dimensional time series vector for each video.

\textbf{Prosody Features (PFs):} We extract the audio from the videos to derive prosody features such as the patterns of rhythm, stress and intonation in speech. We used OpenSMILE \cite{10.1145/2502081.2502224}, a library that helps extract large audio feature spaces in real time. These comprise 1582 features from the INTERSPEECH 2010 Paralinguistic challenge feature set \cite{book} and 382 features from the the INTERSPEECH 2009 Emotion Challenge feature set \cite{inproceedings}. In addition, we use FairPCA \cite{fairpca2018,tantipongpipat2019multi} for finding a low dimensional representation of the features which maintain similar conformity between the multiple groups\footnote{We formed these groups by combining race and gender tags, for e.g : Male Caucasian, Female Caucasian, Male Asian..., etc.} within the dataset.

\subsubsection{Feature Selection and Transformation}
We take three steps to process our features. First, we select a subset of FFs for each social skill based on expert consensus. We only use the selected FFs to build models for a given social skill. Second, we convert the facial time-series features into a vector of aggregate features (AF). Third, we apply a transformation function to convert the aggregate vector for each FF into a single dimensional value FF. We use the same transformation function for all FF feature vectors for a given social skill, which constrains the modeling space further. We describe these three steps in detail now.

\textbf{Feature Selection:} Every social skill links to certain AUs and HPs. For instance, an AU such as an upward lip-curl may signal a smile and indicate ‘Positive Emotion’. We expect a positive correlation - higher the intensity of the AU, higher the ‘Positive Emotion’ score. On the other hand, we do not expect a relationship between say, Lip Puckerer (AU18) or Lip Raiser (AU10) and ‘Positive Emotion’. We recruited five experts with more than 5-10 years of experience in Industrial organizational psychology and diverse ethnicities and gender. For each social skill, they marked whether an FF may signal a positive, a negative or no relationship. FFs were marked as positive indicators, negative indicators, or unrelated, based on expert vote consensus. Only the selected indicators were used to build models for each social skill. No such selection was made on prosody features since they aren't human interpretable.

\textbf{Feature Aggregation:} Typically, researchers average intensities of FFs to aggregate time-series data \cite{chen2017automated}. This does away with multiple rich characteristics of the curve. For instance, a video with bursts of high intensity among a low intensity base, will generate the same value as a constant mid-intensity waveform. Similarly, the variation in the intensity curve may help measure the amount of activity vs. monotony. To capture these effects, we create 11 different aggregate features from the FF time series data. These have been explained in detail in supplementary material. They quantify different characteristics of the intensity curves.

\textbf{Feature Transformation:} The 11 aggregate features for each FF provide richness for machine learning, but exposes the risk of using non-causal correlations. We wished to constrain down the model space of 187 (11*17) features to a lower meaningful dimension. Similar to the use of ‘average intensity’, we learn a common function to aggregate the 11-dimensional vectors across FFs. Rather than one such aggregation function, we create one for every social skill and every type of indicator. The hypothesis is that certain common FF curve properties predict a social skill across, say, positive indicators. This constrains the algorithm to use a common set of curve properties which are predictive of a social skill, than using different properties for each FF. We learn six such functions, covering three social skills \footnote{We remove one social skill out of the initial four. This is explained in sections ahead.} and two indicators each. This may be stated mathematically as follows:
For each social skill,
\begin{center}
\(V_D^{pd} \in\) Pos descriptor set,\, \(V_D^{nd}  \in\) Neg descriptor set 
\[AF_i^{pd} = PosFeatTransformer ( f_1, f_2,........f_n, U^{pd})\]
\[AF_i^{nd} = NegFeatTransformer ( f_1, f_2,........f_n, U^{nd})\]
\(AF_i^{pd}\in V_D^{pd},\, AF_i^{nd} \in V_D^{pd},\, f_i \) aggregate features set\\
\(U_{pd},\,U_{nd}\) unit vectors of dimension equal to no. of FFs in \(V_D^{pd},\, V_D^{pd}\) respectively \\
\end{center}
Let us take the example of positive indicators for ‘Positive Emotion’. We learn a linear function with common weights across selected FFs, but add a different constant for each. We take the 11-dimensional vector as input variables, stacked for all FFs one below the other. Each FF has a separate unit vector, to learn a constant. The social skill score is the output variable and it gets repeated for each stacked FF. We use ridge regression to determine the weights that are most predictive of the given social skill. One may note, all these operations are done on the train set.

We finally have a set off FF features and prosody features for each social skill. We use classical feature selection and machine learning techniques to learn a predictive model.
\subsection{Deep Learning Approach}
Prior research has addressed video processing problems such as video summarization \cite{bilkhu2019attention} and grading video interviews \cite{hemamou2019hirenet} using RNNs, with either LSTMs or GRU cells, and with or without attention. We tried a few different deep-learning approaches and found that a Transformer model with additive attention worked best. We now describe our model structure.

As a first step, we extract 256 dimensional features per-frame from the penultimate layer of a pre-trained CNN model\cite{faceapi}. This model was trained on face images to classify emotion labels (our data has video-wise social skill scores, but not image wise labels). We only use those frames, where a face is detected with a confidence of more than 0.75 and the rest of the frames are imputed with the mean of their surrounding frames. If less than 70\% of the frames in a video have a face in them then we remove the video altogether. We call this 256-dimensional vector for every frame as the face embedding on that frame. When we train the model, the weights that create the embedding are also re-optimized, i.e. fine-tuned.

The number of frames in a video vary. We take the maximum number of frames to be 1200 and perform padding where needed. We then perform dimensionality reduction in the face embedding vector and in the time-space. We used a feedforward layer applied at every time frame to downsample the number of features from 256 to 64. We also add the FF time-series data (described in previous section) to the embeddings to create a 87-dimensional vector. We reduce the number of frames to 239, using a 1D convolution filter with a kernel size of 10 and stride length of 5. We used 87 such filters, each assigned to a particular channel/dimension of the input feature vector. This leads to a 87 by 239 matrix \(X\) per video.

We used multi-head self-attention \cite{vaswani2017attention} on the fixed length sequence \(X\). It contains scaled dot product attention mechanism over the Keys \(K\), Queries \(Q\) and Values \(V\) and compute the representation using softmax as:
\[Q = K = V = X, \quad X\:\textit{represents the input vector matrix}\]
\[Attention (Q, K, V)=softmax( \frac{QK^{T}}{\sqrt{d_k}})V\]
% It contains scaled dot product attention mechanism over the Keys \(K\), Queries \(Q\) and Values \(V\) and compute the representation using softmax as:
% Q inner product with K⊤, and the adjust its value, then use the activation function of softmax to product a new value inner product with the V, and product a new vector.
We then use additive attention \cite{bahdanau} to extract the importance of each feature vector in the sequence. This helps weigh up certain frames, say one representing a lip curl (in case of positive emotion), to amplify their contribution and reduce noise. The additive attention layer first comprises a feedforward layer, followed by a softmax to model the output conditional probability distribution. The final output vector \(h_f\) is the weighted sum (using the probability distribution) of input embeddings across the time dimension. We then concatenate the prosody feature vector \(h_p\) (the output from FairPCA, refer to previous section) with the facial vector \(h_f\) to obtain the final vector. This 151-dimensional vector is then passed onto a feedforward layer (hidden layer of size 64) to get the final prediction. Refer to supplementary material for a detailed explanation.
\section{Experiments \& Results}
\label{headings}
We address the following questions through our experiments:
\begin{itemize}
     \setlength\itemsep{-0.1em}
    \item Do our raters agree to each other and how the different social skills relate with each other?
    \item How do the Expert-Driven (ED) and Deep Learning (DL) models compare on accuracy? Do our models predict ratings with high specificity? 
    \item How do the errors in the model compare within gender and racial groups? Are the models fair and what may be done to mitigate any fairness issues?
    \item Do our predicted social skills predict interview outcomes?
\end{itemize}
\subsection{Video Data and Ratings}
A total of 5845 videos for 810 candidates were collected. All videos marked as 'Video not clear' ($2.2\%$) and cannot be rated ($2.4\%$) by more than 2 raters were removed from the dataset. Most raters judged that videos provided them with sufficient information to rate social skills.

A total of 31 raters were used and every video is rated by at least 5 raters. The inter-rater agreement was measured using pearson coefficient of correlation. The correlation is strong, more than 0.6, for all parameters barring ‘Calmness’ (Refer to supplementary material for Inter-rater agreement details.). This showed that raters across racial and gender groups had a good-level of agreement on the social skill scores of different videos. We averaged the ratings across raters to get the final rating for a video. For a candidate, the ratings across videos were averaged for each social skill, to get the final rating.

We next looked at inter-parameter correlations. These are presented in supplementary material. We find most social skills have a moderate-to-strong correlation with each other ($0.49$-$0.77$). This is expected, since if someone is high in one social skill, generally s/he is high in others too. Confidence and Calmness are very highly correlated by a $r=0.90$. We decided to drop Calmness due to its high correlation with Confidence and also, low inter-rater agreement. A factor analysis of the ratings also confirmed the data represented three significant factors.
\subsection{Model Results}
For training models, we split the candidates into a $70$-$30$ train-test set. We do a random split stratified by ratings. We boot-strap train/test set $20$ times and report average results over runs. Details of all hyperparameters and experimental details are provided in supplementary material.

In Table \ref{model-results}, we report how well our models predict social skills measured by pearson coefficient of correlation. We find that for Confidence and Engagement, both modelling techniques give similar results. For Positive Emotion, DL model gave a much better result, a correlation of $0.65$ as compared to $0.57$ of the ED model. We also split the data by country\footnote{We merged the small sample from other european countries into UK for the analysis.} to see if the correlations remain intact for different countries. The correlations are mostly similar, other than for Confidence, where the DL model shows lower correlation for India.

We also looked at cross-correlations -- how prediction model for one social skill correlated with the expert ratings of a different social skill. This is important to study, when the predicted parameters are correlated. We must check that the prediction model specifically predicts it’s given social skill and doesn’t just signal the shared/common variance with another social skill.

The results are present in Table \ref{inter-model-r}. For the ED model, we find that the model for Positive Emotion predicts Engagement equally well. This shows that the Positive Emotion model is in some sense a proxy model for Engagement, and has picked the shared variance between Positive Emotion and Engagement. In the DL models, we do not come across such an issue.
Finally, we choose the ED models for Confidence and Engagement, since they perform as well as DL models. For Positive Emotion, we choose the DL model for better accuracy and more importantly, specificity. We do a feature analysis and find Positive Emotion is determined mostly by facial expressions, confidence by voice tone and engagement is impacted by both. Details can be found in supplementary material.
\begin{table}
  \caption{Correlation of Expert-Driven and Deep Learning (DL) Models.}
  \label{model-results}
  \centering
  \small
  \begin{tabular}{l l c c c c c c}
    \toprule
     & & \multicolumn{2}{c}{ Positive Emotion} & \multicolumn{2}{c}{Confidence} & \multicolumn{2}{c}{Engagement} \\
    \cmidrule(r){3-8}
    Train Data  & Test Data  & Expert-Driven & DL & Expert-Driven & DL & Expert Driven & DL\\
    \midrule
     All  & All & 0.57 & 0.65 & 0.68 & 0.66 & 0.68 & 0.66 \\
     All  & US-UK & 0.55 & 0.63 & 0.70 & 0.70 & 0.68 & 0.63 \\
     All  & India & 0.56 & 0.65 & 0.67 & 0.63 & 0.67 & 0.65 \\
    \bottomrule
  \end{tabular}
  \vspace{-4mm}
\end{table}
\begin{table}
  \caption{Cross-correlation matrix for Expert-Driven and Deep Learning Models.}
  \label{inter-model-r}
  \centering
  \begin{tabular}{l c c c c c c}
    \toprule
     & \multicolumn{3}{c}{ Expert-Driven }  & \multicolumn{3}{c}{Deep Learning} \\
     \cmidrule(r){2-4}
     \cmidrule(r){5-7}
     & PE* & Confidence & Engagement & PE* & Confidence & Engagement\\
    \midrule
     PE* & 0.57 & 0.38 & 0.48 & 0.65 & 0.44 & 0.55\\
     Confidence& 0.44 & 0.68 & 0.60 & 0.40 & 0.66 & 0.52\\
     Engagement& 0.58 & 0.57 & 0.68 & 0.56 & 0.57 & 0.66\\
    \bottomrule
    \addlinespace[1ex]
    \multicolumn{7}{l}{*PE here represents Positive Emotion}
  \end{tabular}
  \vspace{-4mm}\setlength{\parskip}{0.25\baselineskip}%

\end{table}
% \footnote{We report results on $5$ major racial groups - Caucasians, African-Americans, Asians, Indians and Others.}
\subsection{Fairness Study}
We studied whether our models are fair with regard to gender and racial groups. For each group, we studied two parameters. First, we looked at the mean-difference (mean\_diff) between true and predicted ratings to examine if our models systematically underpredicted or overpredicted scores for any group. Second, we looked at the mean-absolute-error (MAE), to check if the models are less accurate for some groups vs. others.

We calculated the difference in these parameter values, between the group with the highest and lowest value. For example, within racial groups in ‘Positive Emotion’, Afro-Americans had the highest mean-difference ($-0.04$) and Indians\footnote{We have considered sample from India as another race for this analysis. There are many races within India as well, which we plan to study in future work.} have the least mean difference ($0.01$). We take the difference which is $0.05$ (diff\_mean\_diff). Similarly, for MAE, we subtract the values for Others and Asians to get $0.06$ (diff\_MAE). One may refer to supplementary material for detailed numbers.

To understand the impact of these differences, we look at their \textit{effect size} and \textit{significance}. Significance is calculated using 2-sample t-test at $95$\% confidence level. To find effect size, we divide each of the diff\_mean\_diff and diff\_MAE by the standard deviation of the predicted rating parameter (say positive emotion). This tells us, the difference in models is what percentage of standard deviation. Generally, a difference of less than $20$\% of standard deviation is considered small \cite{effectSizeSullivan}.

We find that there is little systematic over/underprediction (signalled by diff\_mean\_diff). All effect sizes are small (less than $20$\%). With regard to accuracy, all differences have small effect sizes, other than Confidence for racial groups which is borderline small ($22.8$\%). It is also interesting to note, that mostly the protected groups (Afro-Americans/Asians, females) have more accurate models (Refer supplementary material).

\begin{table}
  \caption{Effect size and significance of model differences by groups for all social skills.}
  \label{fairness}
  \centering
  \small
  \setlength\tabcolsep{4pt}
  \begin{tabular}{l l l l l l l l l l l l l }
    \toprule
      & \multicolumn{4}{c}{ Positive Emotion } & \multicolumn{4}{c}{Confidence} & \multicolumn{4}{c}{Engagement}  \\
    \cmidrule(r){2-13}
    & \multicolumn{2}{c}{diff\_mean\_diff}& \multicolumn{2}{c}{diff\_MAE}& \multicolumn{2}{c}{diff\_mean\_diff}& \multicolumn{2}{c}{diff\_MAE}& \multicolumn{2}{c}{diff\_mean\_diff}& \multicolumn{2}{c}{diff\_MAE}\\
     \midrule
     type & val & ef-size & val & ef-size & val & ef-size & val &  ef-size & val & ef-size & val & ef-size \\
     \midrule
    Gender& 0.03 & 7.3 & 0.01 & 2.4 & 0.08* & 14.0 & 0.04* & 7.0 & 0.02 & 3.3 & 0.05* & 8.3 \\
    Race & 0.05 & 12.2 & 0.06* & 14.6 & 0.04 & 7.0 & 0.13* & 22.8 & 0.11 & 18.3 & 0.09* & 15\\
    % Gender& 0.03* & 10 & 0.01* & 3.3 & 0.08 & 19.5 & 0.04 & 9.7 & 0.02* & 4.5 & 0.05 & 11 \\
    % Skin-color & 0.03*** & 10 & 0.03* & 10  &  0.12 & 29.2 & 0.08 & 19.5 & 0.12 & 27  & 0.03*** & 6.8  \\
    % Race & 0.05* & 16.67 & 0.06 & 20 & 0.04* & 9.7 & 0.13 & 31.7 & 0.11* & 25 & 0.09 & 20\\
    \bottomrule
    \addlinespace[1ex]
    \multicolumn{13}{l}{\textsuperscript{*}$p<0.05$, \textit{val} represents the maximum absolute difference }
  \end{tabular}
  \vspace{-4mm}
\end{table}
We also investigated whether the models will lead to fair selection if used with a cut-score. We applied a cut-score that eliminated the bottom one-third candidates\footnote{Companies generally use these tools to eliminate the bottom performers and do further rounds of interviews with the rest.}. We find that the Disparate Impact\cite{feldman2015certifying} for each sensitive attribute is well between $80\%$-$120\%$. Please refer to supplementary material for more details.

We present the results in Table \ref{fairness}. These are very encouraging results. Only one effect sizes is not small, but on the borderline. This could be mitigated in couple of ways. First, one could investigate further machine learning techniques to induce fairness in the models. Techniques for data-transformation \cite{NIPS2017_optimizedpreprocessing}, fair feature engineering \cite{tantipongpipat2019multi,fairpca2018, GrgicHlaca2018BeyondDF} and post-processing \cite{NIPS2016_6374_post} may be used. Secondly, certain standard practices in assessment may be used. For instance, consider that a company wants to shortlist candidates based on the Confidence score. They estimate a cut-score based on the trait requirement for the job. To ensure fairness, they could lower the estimated cut-score for the group with the higher error. If companies do not want to use different cut-scores for different groups, they can uniformly lower the cut-score, allowing passage to candidate who are impacted by model accuracy. The hypothesis is that the non-worthy candidates will be out-selected in further expert rounds. Our larger recommendation is that these models provide very promising results with regard to fairness, but must be used with oversight and caution.

\subsection{Validation Study}
We tested whether our scores are predictive of interview success for three different companies across India and China. The companies were hiring for a product engineering role. All the applicants took our asynchronous interview. For the first two companies, company personnel did an independent interview for all the applicants and made offers. In case of the third, the company personnel rated the videos on a 3-point scale on hireability. Company personnel did not have access to the social skill scores in any of the three studies. All correlations (ranging $0.12$-$0.52$) between social skill scores and interview outcomes are significant, other than one. We also did a regression using the three scores and the regression $r$ range from $0.29$-$0.56$. These correlations are similar or better than the correlation of personality scores with interview outcomes, that range $0.20$-$0.35$ \cite{cook2000relation,caldwell1998personality}. Also, we must consider that social skills are not expected to explain all the variance of interview outcomes, since interviewers also assess other parameters such as domain skills. Refer to supplementary material for more details.
\section{Conclusion}
In this work, we use a multi-country and multi-racial dataset for building models to grade video interviews and study fairness with regard to gender and race. We measure externally expressed behaviour which we call social skills. We adopt several best practices in the rating process to arrive at a true measure of social skills. We build two models, first one using expert-verified features, while the second is a complex state-of-the-art transformer model with additive attention. We find that the simpler model performs similar to the complex model for two of the parameters. The model also generalizes well across candidates from different countries. We also study the errors of video grading models with regard to gender and race. We find most effect sizes to be small and the results to be very encouraging. We finally verify that the social scores are indeed predictive of interview outcomes. In future work, we plan to do a more detailed analysis of model fairness and also, experiment with more methods to mitigate biases in models. We also plan to test the validity of scores against on-job performance.
\section*{Broader Impact}
This study presents the first examination of fairness in the scoring of video interviews. This work should help companies assess candidates fairly and recruit a more diverse workforce by minimizing bias when evaluating job candidates, particularly candidates from protected classes. We consider ways to construct models that use causal features. We also examine whether the grading models exhibit bias by race or gender. We urge practitioners to develop and use these models with utmost care, applying the concepts developed in this paper. We hope this work will help encourage and develop further studies in fairness in video processing.
\bibliographystyle{plain}
\bibliography{paper.bib}
\end{document}

\maketitle
\section{Dataset and Ratings}
\subsection{Demographics of candidates and raters}
\def \hfillx {\hspace*{-\textwidth} \hfill}
\begin{table}[h]
    \small
    \begin{minipage}{0.6\textwidth}
        \caption{Demographics of candidates participating in the exercise.}
        \label{candidates}
        \centering
        \setlength\tabcolsep{3pt}
        \begin{tabular}{l c c}
        \toprule
        Category & Candidates (count) & Candidates (\%) \\
        \midrule
        India & 411 & 50.7\\
        US & 156 & 19.2 \\
        UK & 182 & 22.5 \\  
        Other Countries (Europe) & 61 & 7.5 \\
        \midrule
        Caucasian & 189 & 23.3 \\
        Afro-American & 61 & 7.5\\
        Asian & 71 &8.8\\
        Others & 78 & 9.6\\
        Indians & 411 & 50.7\\
        \midrule
        Male & 424 & 52.3 \\
        Female & 386 & 47.7 \\
        \midrule
        18-20 yrs & 30 & 3.7\\
        20-30 yrs & 565 & 69.7\\
        30-40 yrs & 126 & 15.6\\
        >40 yrs & 89 & 11.0\\
        \bottomrule
        \end{tabular}
    \end{minipage}
    \hfillx
    \begin{minipage}{0.4\textwidth}
        \centering
        \setlength\tabcolsep{3pt}
        \label{raters}
        \caption{Demographics of raters.}
        \begin{tabular}{l c}
        \toprule
        Category & Count\\
        \midrule
        Caucasian & 9\\
        Asian & 9 \\
        African-American & 9 \\
        Others & 4\\
        \midrule
        Male & 17\\
        Female & 14\\
        \bottomrule
      \end{tabular}
    \end{minipage}
\end{table}

\subsection{Sample Questions and Rubric}
\begin{table}[h]
  \scriptsize
  \centering
  \caption{Examples of questions asked in the interview.}
  \label{question}
  \begin{tabular}{l l}
    \toprule
    Type & Question\\
    \midrule
    Competency-based & Tell us a time when you got some useful feedback to improve yourself? Probably, you made a mistake or did not put\\ 
    & required effort in completing a task or project?What did you do based on the feedback?\\
    \midrule
    Domain (Sales) & You work as a sales manager for a marketing agency. During a contract renewal meeting, one of your major clients says \\
    & that your competitor has offered them services at a lower price. How will you handle the situation?\\ 
    \bottomrule
  \end{tabular}
\end{table}
\begin{table}[h]
  \scriptsize
  \centering
    \caption{Sample Rubric for the social skill - Engagement}
  \label{rubric}
  \begin{tabular}{l l}
    \toprule
    Level & Description\\
    \midrule
    NA & \textbf{Can’t say based on Video}\\
    \midrule
    4 & The candidate is expressive. She/he shows strong interest in the content being delivered and speaks in a manner that is engaging \\ 
    & to the listener. She/he demonstrates excitement when appropriate.\\ 
    \midrule
    3 & \textit{ In between these two levels}\\
    \midrule
    2 & The candidate is expressive at times and shows some interest in the content being delivered. S/he is moderately engaging and could \\
    & do better.\\
    \midrule
    1 & \textit{ In between these two levels}\\
    \midrule
    0 & The candidate lacks expressiveness. Her/his speech is monotonous. She/he seems uninvolved and uninterested in the content\\
     & s/he is delivering. \\
     \midrule
    -1 & Video not clear \\
    \bottomrule
  \end{tabular}
\end{table}
The interview had five behavorial and two domain knowledge questions (See samples in Table \ref{question}). Domain questions came from multiple areas such as technology, banking, sales and accounting.

We defined a rubric for each social skill. As an example, one may refer to rubric of social skill, engagement, in Table \ref{rubric}.

\section{Methods}
\subsection{Related Work}
Researchers have used four kind of features for grading video interviews- facial expression/action unit intensities/emotions \cite{hemamou2019hirenet,naim2016automated,chen2017automated,hireme}; prosody features to judge voice tone \cite{naim2016automated,hireme}; speech likelihood \cite{chen2017automated} and fluency \cite{chen2017automated,hireme} features based on ASR (automatic speech recognition) alignments; and natural-language features derived from the automatically transcribed text. In \cite{hireme}, they extract features from both the applicant and the interviewer. Most approaches \cite{naim2016automated,hireme} simply average the features across time to aggregate them, use standard feature selection/dimensionality reduction techniques and learn models using classical machine learning techniques such as SVMs and Random Forest. In \cite{hemamou2019hirenet}, authors proposed a hierarchical deep-learning based attention model to classify a hire/not-hire. They use Word2Vec embeddings from text, eGeMAPS features for audio from Opensmile \cite{10.1145/2502081.2502224} and facial action unit and relative head rotations from each video, which are then processed by a GRU with additive attention.
\subsection{Expert-Driven Approach}
Definitions of the 11 different aggregate features derived from the FF time series data.\\
\begin{table}[h]
    \caption{List of aggregate features and description.}
    \label{ml-feats}
    \scriptsize
    \centering
    \begin{tabular}{ l l }
    \toprule
    Feature Type &  Description  \\
    \midrule
     num\_intensity\_rise  & Number of instances where intensity rise between all closest  minima and maxima pairs is above a threshold \\
     & normalized by the duration of the video.\\
     num\_intensity\_drop  & Number of instances where intensity fall between all closest minima and maxima pairs is above a threshold\\
     & normalized by the duration of the video.\\
     mean\_max\_intensities & Mean of all local maximas obtained from time series intensity plot.\\
     mean\_min\_intensities & Mean of all local minimas obtained from time series intensity plot.  \\
     average\_singular\_change & Average change of intensity between 2 consecutive timestamps\\
     & timestamps in the whole video.\\ 
     longest\_pos\_run & Length of longest pos trend normalized by the video duration.\\
     longest\_neg\_run & Length of longest neg trend normalized by the video duration. \\
     stdev & Standard deviation  \\
     max\_intensity\_5 & Average of the top 5 local maximas  \\ 
     num\_maximas & Number of local maximas normalized by the duration of the video\\
     num\_minimas & Number of local minimas normalized by the duration of the video\\
    \bottomrule
  \end{tabular}
\end{table}
\subsection{Deep Learning Approach}
\begin{figure}[h]
    \centering
    \fbox{
    \includegraphics[width=0.95\textwidth]{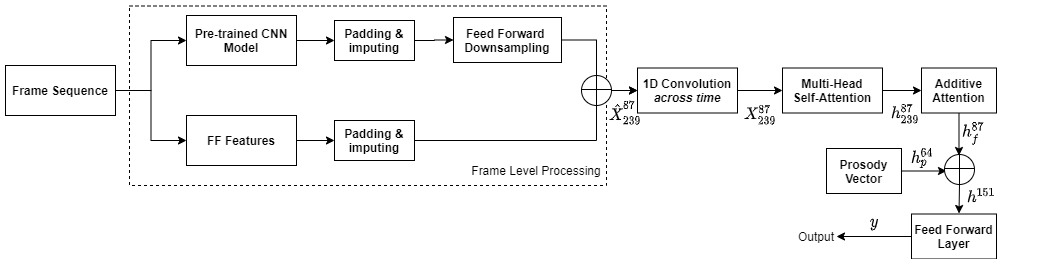}
    }
  \caption{Deep Learning model architecture flow}
  \label{dlflow}
\end{figure}
We describe the deep learning model in detail. All these steps are illustrated in Fig \ref{dlflow}.
As a first step, we extract 256 dimensional features per-frame from the penultimate layer of a pre-trained CNN model\cite{faceapi}. This model employs depthwise separable convolutions and densely connected blocks. This model was trained on face images to classify emotion labels (our data has video-wise social skill scores, but not image wise labels). When we train the model, the weights that create the embedding are also re-optimized, i.e. fine-tuned.
We now describe the pre-processing steps taken:

\textbf{Imputing:} We only use those frames, where a face is detected with a confidence of more than 0.75 and the rest of the frames are imputed with the mean of their surrounding frames. If less than 70\% of the frames in a video have a face in them then we remove the video altogether. We call this 256-dimensional vector for every frame as the face embedding on that frame. 

\textbf{Padding:} The number of frames in a video vary. We take the maximum number of frames to be 1200 and perform padding where needed. 

We then perform dimensionality reduction in the face embedding vector and in the time-space. We used a feedforward layer applied at every time frame to downsample the number of features from 256 to 64. We also add the FF time-series data (described in main paper) to the embeddings to create a 87-dimensional vector. We reduce the number of frames to 239, using a 1D convolution filter with a kernel size of 10 and stride length of 5. We used 87 such filters, each assigned to a particular channel/dimension of the input feature vector. This leads to a 87 by 239 matrix \(X\) per video.
\[Input\,\hat{X}=\{ \hat{x}_{0}^{87}, \hat{x}_{1}^{87},.........\hat{x}_{1198}^{87}, \hat{x}_{1199}^{87}  \}\]
\[Output\,X=\{ x_{0}^{87}, x_{1}^{87},.........x_{237}^{87},  x_{238}^{87}  \}\]

We used multi-head self-attention \cite{vaswani2017attention} on the fixed length sequence \(X\). It contains scaled dot product attention mechanism over the Keys \(K\), Queries \(Q\) and Values \(V\) and compute the representation using softmax as:
\[Q = K = V = X, \quad X\:\textit{represents the input vector matrix}\]
\[Attention (Q, K, V)=softmax( \frac{QK^{T}}{\sqrt{d_k}})V\]
% It contains scaled dot product attention mechanism over the Keys \(K\), Queries \(Q\) and Values \(V\) and compute the representation using softmax as:
% Q inner product with K⊤, and the adjust its value, then use the activation function of softmax to product a new value inner product with the V, and product a new vector.
We then use additive attention \cite{bahdanau} to extract the importance of each feature vector in the sequence. This helps weigh up certain frames, say one representing a lip curl (in case of positive emotion), to amplify their contribution and reduce noise. The additive attention layer first comprises a feedforward layer, followed by a softmax to model the output conditional probability distribution. 
\[H= [h_0,h_1,h_2......h_{238}]\]
\[c= tanh(WH + b)\]
\[\alpha_j^i= exp(c_j^i)  /\sum\limits_{t}exp(c_t^i)\]
\[h_f= \sum \alpha_t.h_t\]
\(W \in R^{239X239} \) is a weight matrix, \(b\in R^{239}\) bias-vector and \(h_t \in R^{87}\) denotes the output returned from the self-attention layer at time t.

The final output vector \(h_f\) is the weighted sum (using the probability distribution) of input embeddings across the time dimension. We then concatenate the prosody feature vector \(h_p\) (the output from FairPCA, refer to main paper) with the facial vector \(h_f\) to obtain the final vector. This 151-dimensional vector is then passed onto a feedforward layer (hidden layer of size 64) to get the final prediction. 
\[h=[h_f,h_p]\]
\[y=Wh+b\]
% Firstly, basic preprocessing of padding, imputing and downsampling of face embedding was performed. After that, we applied 1D convolution layer across time to reduce the time dimension space of data. This reduces the complexity of the model and not effect the data loss as adjacent frame contain correlated features. 87 1D convolution filter of kernel size 10 and stride length of 5 were used. This leads to a 87 by 239 matrix \(X\) per video.

% We applied multi-head self-attention on the \textit{ X}. 
% \[Q = K = V = X, \quad X\:\textit{represents the input vector matrix}\]
% \[Attention (Q, K, V)=softmax( \frac{QK^{T}}{\sqrt{d_k}})V\]

% We then applied additive attention layer on output returned from the self-attention layer at time t. 

% \[Input\, Vector=\{ f_{0}^{87}, f_{1}^{87},.........f_{1198}^{87}, f_{1199}^{87}  \}\]
% \[Output\, Vector=\{ \acute{f}_{0}^{87}, \acute{f}_{1}^{87},.........\acute{f}_{237}^{87},  \acute{f}_{238}^{87}  \}\]

% \[K_i = W_KK\,,\,V_i=W_VV\,,\,Q_i = W_QQ\] 
% \[Attention (Q, K, V)=softmax( Q_iK_iT/ d_i) V_i\]
% \[where\,W_K, W_V, W_Q\,are\,the\,corresponding\,transformation\,matrices\]
\section{Experiment \& Results}
\subsection{Video Data and Ratings}

\begin{table}[h]
    \caption{Inter-rater agreement (IRA) for each of the social skill.}
      \label{agreement}
      \small
      \centering
      \begin{tabular}{l c}
        \toprule
        Parameter & IRA \\
        \midrule
         Positive Emotion  & 0.62\\
         Calmness  &  0.56\\
         Confidence  & 0.62\\
         Engagement & 0.67 \\
         Average Social Skill & 0.69 \\
        \bottomrule
      \end{tabular}
\end{table}

\begin{table}
    \caption{Inter-parameter correlations in the final ratings.}
      \label{actualr}
      \setlength\tabcolsep{5pt}
      \small
      \centering
      \begin{tabular}{l c c c c c}
        \toprule
        All Data & PE & Calmness & Confidence & Engagement  \\
        \midrule
         PE  & 1.00 & 0.49 & 0.54 & 0.77 \\
         Calmness & 0.49 & 1.00 & 0.90 & 0.61\\
         Confidence & 0.54 & 0.90 & 1.00 & 0.68\\
         Engagement  &0.77 & 0.61 & 0.68 & 1.00 \\
        \bottomrule
        \addlinespace[1ex]
        \multicolumn{3}{l}{\textsuperscript{*} PE here represents Positive Emotion}
      \end{tabular}
\end{table}

The discussed inter-rater agreement and the inter-parameter correlations have been reported in Table \ref{agreement} and \ref{actualr}.

\subsection{Experimental parameters}
% For training models, we split the candidates into a 70-30 train-test set. We do a random split stratified by ratings. We bootstrap train-test set 20 times and report the average results.\\
For the Expert-Driven model, we experimented with linear regression with L-1 (LASSO) and L-2 regularization (Ridge), and Random Forests. For Ridge, the optimal coefficient \(\alpha\) is determined by varying it between 1 to 1000 and selecting the value with the best cross-validation correlation. For LASSO \cite{lasso} (\(\alpha\)= 1) we varied \(\lambda\) from 0 to 4. For Random forests \cite{randomforest}, we varied the number of estimators from 15 to 100. In all techniques, the model which gave the best cross-validation correlation was selected. Ridge gave the best results for all social skills. We report its results in the main paper.\\
For the Deep Learning model, we used a Nesterov Adam optimizer (NAdam)\cite{dozat2016incorporating} which uses Nesterov accelerated gradient instead of momentum in Adam optimizer. We used mean square error as the loss function for the optimizer. We used L-2 regularization (alpha ~ 0.0005) and dropout (0.4-0.5) to address overtraining. A mini batch size of 8 to 16 is used, with an Exponential Linear Unit (ELU) \cite{shah2016deepelu} activation function deployed after each layer. Since we used face embeddings from a pre-trained model, we did not have any scale invariance issues on the facial data. BatchNormalization\cite{ioffe2015batchnormalization} is used to normalize each batch before feeding to the subsequent layer. This reduces the amount of change in the hidden unit values (covariance shift) and speeds up the training process. We used an early stopping technique by using validation loss and correlation to evaluate the number of epochs of training. This further helps in avoiding overfitting to the training set.

\subsection{Feature Importance}
There is considerable interest in the human behavior and psychology community regarding the relative contribution of voice vs. facial expressions to different social skills.  We use a model inspection technique, called permutation importance\cite{altmann2010permutation}, to determine feature importance. Here we randomly shuffle the feature values of one set of features at a time and report the effect on test correlation. In Table \ref{feats-imp}, we report the feature importance of facial and prosody components in our final models. Largely, we find Positive Emotion is determined by facial expressions, confidence by voice tone and engagement is impacted by both. This matches our intuition: positive emotion is signalled by a smile, confidence is signalled by a confident, non-stuttering voice and engagement has components of both - expressive face and voice modulation.
\begin{table}[h]
 \small
  \caption{Feature Importance of facial and prosody components in Expert-Driven and Deep Learning models.}
  \label{feats-imp}
  \centering
  \begin{tabular}{l c c c c}
    \toprule
     & \multicolumn{2}{c}{Expert-Driven} & \multicolumn{2}{c}{Deep Learning}  \\
    \cmidrule(r){2-5}
    Social Skills & facial & prosody & facial & prosody \\
    \midrule
    Positive-emotion & 0.28 & 0.12 & 0.46 & 0.07 \\
    Confidence & 0.03 & 0.26 & 0.07 & 0.48 \\
    Engagement & 0.20 & 0.20 & 0.17 & 0.11 \\
    \bottomrule
  \end{tabular}
  \vspace{-6mm}
\end{table}

\section{Fairness Study}
There has been a lot of research around the fairness of machine learning algorithms, especially where there are direct implications of these algorithms be it criminal risk assessments \cite{criminalRisk} or credit-scoring \cite{creditscoring} or decision making in a partly automated workflow \cite{waters:iaai13}. \cite{agarwal2019fair} propose schemes for fair regression under statistical parity and bounded group-loss, in \cite{berk2017convex} authors look at notions of individual and group fairness. \cite{ETSFairnessVideo} investigate the fairness of human ratings when scoring video interviews, \cite{tedtalk} detect and mitigate bias under the notion of disparate impact.
\subsection{Regression Metrics}
In Table \ref{fairness}, we report the model errors i.e mean\_diff and MAE for each category of the group.
\begin{table}[h]
  \caption{Model errors (mean\_diff and MAE) of each group for all the social skills.}
  \label{fairness}
  \centering
  \small
  \setlength\tabcolsep{5pt}
  \begin{tabular}{l c c c c c c}
    \toprule
      & \multicolumn{2}{c}{ Positive Emotion } & \multicolumn{2}{c}{Confidence} & \multicolumn{2}{c}{Engagement}  \\
    \cmidrule(r){2-7}
    & mean\_diff & MAE & mean\_diff & MAE & mean\_diff & MAE \\
    \midrule
    Male & 0.00 & 0.28 & -0.04 & 0.43 & -0.09 & 0.49\\
    Female & -0.03 & 0.27 & -0.12 & 0.39 & -0.11 & 0.44\\
    \midrule
    Caucasians & -0.03 & 0.29 & -0.07 & 0.38 & -0.04 & 0.49\\
    Asian & -0.02 & 0.24 & -0.07 & 0.38 & -0.12 & 0.45\\
    African-Americans & -0.04 & 0.28 & -0.10 & 0.32 & -0.15 & 0.40\\
    Indian & 0.01 & 0.27 & -0.07 & 0.45 & -0.11 & 0.47\\
    Others & -0.02 & 0.30 & -0.06 & 0.39 & -0.11 & 0.46\\
    \bottomrule
  \end{tabular}
  \vspace{-4mm}
\end{table}

\subsection{Classification Metrics}

We also investigated whether the models will lead to fair selection if used with a cut-score. We applied a cut-score that eliminated the bottom one-third candidates\footnote{Companies generally use these tools to eliminate the bottom performers and do further rounds of interviews with the rest.}. We evaluate our models on three notions of group fairness - Equalized Odds \cite{hardt2016equality}, Equal Opportunity \cite{hardt2016equality} and Demographic parity \cite{verma2018fairness}. We also calculate Disparate Impact \cite{feldman2015certifying}, which compares the proportion of candidates selected from protected vs. privileged groups. We consider females, Afro-Americans, Asians, Indians, and Others as protected groups. Males and Caucasians are considered privileged.

We present the results in Table \ref{classfairness}. Disparate Impact (DI) and Equalized Odds (EO) are determined using both the actual and predicted ratings. Equal Opportunity and Demographic Parity signify the accuracy of selections and overall accuracy respectively. The values of these metrics don't differ much between protected and privileged groups. Wherever there is a considerable difference, protected groups are favored.  We find that the DI for each sensitive attribute is well between $80\%$-$120\%$. This exhibits that the amount of unfairness present in the original dataset and the model predictions w.r.t gender and racial groups is well within the limits.
% We present the results in Table \ref{classfairness}. Disparate Impact (DI) and Equalized Odds (EO) are determined using both the actual ratings and predicted model scores. We find that the DI for each sensitive attribute is well between $80\%$-$120\%$. This exhibits that the amount of unfairness present in the original dataset and the model predictions w.r.t gender and racial groups is well within the limits. Equal Opportunity and Demographic Parity signify the accuracy of selections and overall accuracy respectively. The values for in both the metrics don't differ much between protected and privileged groups, wherever there was a difference it was generally more for protected groups. Thus not propagating any perceived bias against protected groups.

% \textbf{Demographic Parity}
% C is independent of A: P₀ [C = c] = P₁ [C = c] ∀ c ∈ {0,1}
% \textbf{Equalized odds:}
% P₀ [C = r | Y = y] = P₁ [C = r | Y = y] ∀ r, y
%  The limitation of this weaker notion is that we can trade false positive rate of one group for false negative rate of another group. Such trade is not desirable sometimes(e.g. trade rejecting(C=0) qualified applicants(Y=1) from group1 (A=0) for accepting(C=1) unqualified people(Y=0) from group2 (A=1) )
% \textbf{Equal Opportunity:}
%  P₀ [C = 1| Y = 1] = P₁ [C = 1| Y = 1]
% which is called Equality of Opportunity.
% We find that the Disparate Impact \cite{feldman2015certifying} for each sensitive attribute is well between $80\%$-$120\%$.

\begin{table}[h]
  \caption{Equalized Odds (EO), Equal Opportunity, Demographic Parity and Disparate Impact (DI) evaluated for all the social skills pertaining to gender and racial groups.}
  \label{classfairness}
  \centering
  \small
  \setlength\tabcolsep{3pt}
  \begin{tabular}{l c c c c c c}
    \toprule
    Category & EO (Actual) & EO & Equal Opportunity & Demographic Parity & DI(Actual) & DI  \\
     \midrule
    Male & 0.62 & 0.69 & 0.54 & 0.78 & & \\
    Female & 0.72 & 0.81 & 0.68 & 0.82 & 117\% & 116\% \\
    % fair & 0.75 & 0.72 & 0.64 & 0.81 & 1.00 & 0.80\\
    % medium & 0.73 & 0.68 & 0.61 & 0.81 &  & \\
    % dark & 0.75 & 0.58 & 0.54 & 0.76 &  & \\
    \midrule
    Caucasian & 0.73 & 0.78 & 0.66 & 0.80 &  & \\
    Asian & 0.71 & 0.75 & 0.65 & 0.84 & 97\% & 96\%\\
    Afro-Americans & 0.67 & 0.84 & 0.66 & 0.82  & 91\% & 108\% \\
    Indians & 0.62 & 0.71 & 0.55 & 0.77 & 95\% & 90\%\\
    Others & 0.70 &  0.70 & 0.63 & 0.86 & 84\% & 92\% \\
    \bottomrule
  \end{tabular}
\end{table}

\begin{table}[h]
 \small
  \caption{Correlations of social scores with interview performance; Sample sizes for each study given besides company}
  \label{ValidationStudy}
  \centering
  \begin{tabular}{l l l l}
    \toprule
    & Company A (59) & Company B (226) & Company C (56) \\
    \midrule
    Positive Emotion & 0.10  & 0.12  & 0.45* \\
    Confidence & 0.32*  & 0.18* & 0.52*  \\
    Engagement & 0.38*  &  0.24*  & 0.47* \\
    Regression results & 0.40 & 0.29 & 0.56 \\ 
    \bottomrule
    \addlinespace[1ex]
    \multicolumn{4}{l}{\textsuperscript{*}$p<0.05$}
  \end{tabular}
\end{table}
\section{Validation Study}
We tested whether our scores are predictive of interview success for three different companies across India and China. The companies were hiring for a product engineering role. All the applicants took our asynchronous interview and were automatically scored. For the first two companies, company personnel did an independent interview of all the applicants and hired some of them (1-hired, 0-not hired). In case of the third company, the company personnel rated the videos on a 3-point scale ('Can be hired', 'Maybe hired' and 'Cannot be hired'). Company personnel did not have access to the automated social skill scores in any of the three studies, at the time of making decisions. Table \ref{ValidationStudy} shows the sample sizes and the correlation (with significance) of the social skill scores with the hiring status. We also do a linear regression with the three scores to predict the hiring status.\\
All correlations are significant, other than one. The total regression coefficients range from 0.29-0.56. This is similar or better to the correlation of personality scores with interview outcomes, that range $0.20$-$0.35$ \cite{cook2000relation,caldwell1998personality}. Also, we must consider that social skills are not expected to explain all the variance of interview outcomes, since interviewers also assess other parameters such as domain skills and language skills.\\
These studies validate that our social skill scores are indeed predictive of interview outcomes. In future work, we plan to test the validity of these scores against on-job-performance.

\bibliographystyle{plain}
\bibliography{supplement.bib}